%% file: arxiv_main.tex
\pdfoutput=1

\documentclass[11pt]{article}

\usepackage{ACL2023}
\usepackage{times}
\usepackage{latexsym}

\usepackage{amsfonts} 
\usepackage{multirow}
\usepackage{amsmath}
\usepackage{hyperref}

\usepackage{graphicx}		
\usepackage{subfigure}
\usepackage{booktabs}
\usepackage{subfigure}

\usepackage{algorithm}
\usepackage{algpseudocode}
\usepackage{longtable}

\usepackage[T1]{fontenc}

\usepackage[utf8]{inputenc}

\usepackage{microtype}

\usepackage{inconsolata}
\usepackage{soul}

\title{AutoPRM: Automating Procedural Supervision for Multi-Step Reasoning via Controllable Question Decomposition}

\author{Zhaorun Chen\thanks{$^{*}$Work was done during Zhaorun Chen’s remote internship at UNC.}$^{\phantom{*}1}$, Zhuokai Zhao$^{1}$, Zhihong Zhu$^{2}$, Ruiqi Zhang$^{3}$\\ \textbf{Xiang Li$^{4}$, Bhiksha Raj$^{4}$, Huaxiu Yao$^{5}$} \\
$^{1}$University of Chicago, $^{2}$Peking University,$^{3}$University of California, Berkeley\\ $^{4}$Carnegie Mellon University, $^{5}$UNC-Chapel Hill\\
\texttt{billchandler0226@gmail.com}, \texttt{huaxiu@cs.unc.edu}}

\begin{document}
\maketitle
\begin{abstract}

Recent advancements in large language models (LLMs) have shown promise in 
multi-step reasoning tasks, 
yet their reliance on extensive manual labeling to provide procedural feedback remains a significant impediment.
To address this challenge, in this paper, we propose a novel self-supervised framework \textbf{AutoPRM} that efficiently
enhances the fine-tuning of LLMs for intricate reasoning challenges. 
Specifically, \textbf{AutoPRM} first decomposes complex problems into more manageable subquestions 
with a controllable granularity switch, 
then sequentially apply reinforcement learning to iteratively improve the subquestion solver. 
Additionally, we propose context-guided-decoding to avoid reward tampering and guide the subquestion solver towards the solution of the holistic problem.
Extensive experiments show that \textbf{AutoPRM} significantly 
improves performance on mathematical and commonsense reasoning tasks over SOTA. 
More encouragingly, \textbf{AutoPRM} can be easily integrated with other orthogonal reasoning 
pipelines.
%
%
\end{abstract}

\input{content/introduction}
\input{content/method}
\input{content/experiment}

\input{content/related_work}
\input{content/conclusion}
\newpage
\section*{Limitations}
While AutoPRM significantly improves reasoning accuracy on arithmetic tasks, we observe 
marginal improvement on StrategyQA~\cite{geva2021did}, regardless of fine-tuning or 
decoding methods, as showed in Table~\ref{tab:limitation_strategyQA}. 
However, instead of an indicator for poor reasoning capability, we suspect such result 
could be due to the model knowledge gap, as discussed in~\cite{petroni2019language}, since StrategyQA highly depends on factual truthfulness. In this case, continuing to enhance reasoning reliability might provide very limited performance gain.
And we believe increasing the scale of model parameters and training data is the key factor to further improve accuracy on 
such knowledge-dependent reasoning tasks~\cite{anil2023palm}.
%


\section*{Acknowledgements}
We thank Google Cloud Research Credits program for supporting our computing needs.

\bibliography{custom}
\bibliographystyle{acl_natbib}

\onecolumn
\appendix
\section{Appendix}
\label{sec:appendix}
\subsection{Detailed Training Procedures}
\label{sec:appendix_hyperparameters}

The complete procedures and hyperparameters (Table~\ref{tab:hyperparameter_poar}) for fine-tuning AutoPRM are detailed in this section.
All models including SFT and RL-fine-tuned models are fully fine-tuned based on LLaMA-2-7B~\cite{touvron2023llama}.
Specifically, the RL via expert iteration process is iterated five epochs, with the
best model being selected based on its performance in final-answer error on the validation set.
All model training was conducted using Huggingface Library~\cite{wolf2020transformers}.

\begin{table}[h]
\centering
\begin{tabular}{l|c}
\hline
\textbf{Parameters} & \textbf{Value} \\ \hline
learning rate (SFT) & 1e-4  \\ \hline
learning rate (RL)  & 5e-5  \\ \hline
learning rate scheduler & cosine  \\ \hline
batch size & 32 \\ \hline
weight decay & 0.05\\ \hline
warmup steps & 100\\ \hline
\end{tabular}
\caption{AutoPRM hyperparameter settings}
\label{tab:hyperparameter_poar}
\end{table}

\newpage
\subsection{Data Retrieval Prompts}\label{sec:appendix_subQG}
\subsubsection{Subquestion Collection (SQC) Prompts}
\begin{longtable}{p{15.7cm}}
\toprule
    Here is is a subsolution to a grade school math question. You should first (1) rephrase information (e.g. numbers, conditions) from the context necessary to reconstruct the subsolution, (2) delete redundant information not used in the subsolution, (3) ask a subquestion based on this subsolution. Please make sure you include all the necessary information in the subsolution. 
    \\
    \\
    Example 1:
    \\
    \\
    \textbf{Context}: Janet\u2019s ducks lay 16 eggs per day. She eats three for breakfast every morning and bakes muffins for her friends every day with four. She sells the remainder at the farmers' market daily for \$2 per fresh duck egg. 
    \\
    \\
    \textbf{Subsolution}: Janet sells 16 - 3 - 4 = <<16-3-4=9>>9 duck eggs a day. 
    \textbf{Subquestion}: Janet\u2019s ducks lay 16 eggs per day. She eats three herself and bakes muffins with four. She sells the remainder to the market. How many duck eggs does Janet sell a day?
    \\
    \\
    Example 2:
    \\
    \\
    \textbf{Context}: Every day, Wendi feeds each of her chickens three cups of mixed chicken feed, containing seeds, mealworms and vegetables to help keep them healthy.  She gives the chickens their feed in three separate meals. In the morning, she gives her flock of chickens 15 cups of feed.  In the afternoon, she gives her chickens another 25 cups of feed.
    \\
    \\
    \textbf{Subsolution}: If each chicken eats 3 cups of feed per day, then for 20 chickens they would need 3*20=<<3*20=60>>60 cups of feed per day. 
    \textbf{Subquestion}: Each day Wendi feeds each of her chickens three cups of mixed chicken feed. How many cups of feed do 20 chickens need per day? 
    \\
    \\
    \textbf{Context}:  \\
    \textbf{Subsolution}: \\
    \textbf{Subquestion}: \\
\bottomrule
\caption{The prompt input to GPT-3.5 for subq-question collection on GSM8K dataset} \label{tab:prompt.GSM}
\end{longtable}

\newpage
\begin{longtable}{p{15.7cm}}
\toprule
    Here is a solution of multiple steps to a grade school math question. Please break the question down into several intermediate questions that ask the result of each intermediate step of the solution. You should provide the context of the original question first, then provide the intermediate questions. For example: 
    \\
    \\
    \textbf{Original Question}: Lana is brewing cups of tea for her friends. She has 27 cups, and she divides these into 3 rows. In each row, she creates equal amounts of chamomile and mint tea cups. She then uses the remaining cups to brew a total of 15 cups of cinnamon tea. How many cups of mint tea are in each row? 
    \\
    \\
    \textbf{Original Solution}: If there are 15 cups of cinnamon tea, then there are a total of 27 - 15 = <<27-15=12>> 12  cups of chamomile or mint tea.\\
    As there are equal amounts of chamomile tea and mint tea, there is a total of 12 cups / 2 = <<12/2=6>>6 cups of mint tea.\\
    Dividing these into rows shows that each row holds 6 / 3 = <<6/3=2>>2 cups of mint tea. \\
    \#\#\#\# 2 \\
    The answer is: 2 
    \\ 
    \\ 
    \textbf{Break Down}: \\
    \textbf{Context}: Lana is brewing cups of tea for her friends. She has 27 cups, and she divides these into 3 rows. In each row, she creates equal amounts of chamomile and mint tea cups. She then uses the remaining cups to brew a total of 15 cups of cinnamon tea.  \\
    \\
    \textbf{Intermediate Question 1}: How many cups of chamomile or mint tea are there? \\
    \textbf{Intermediate Solution 1}: If there are 15 cups of cinnamon tea, then there are a total of 27 - 15 = <<27-15=12>>12 cups of chamomile or mint tea. \\
    \#\#\#\# 12 \\
    The answer is: 12 \\
    \\
    \textbf{Intermediate Question 2}: How many cups of mint tea are there? 
    \textbf{Intermediate Solution 2}: If there are 15 cups of cinnamon tea, then there are a total of 27 - 15 = <<27-15=12>>12 cups of chamomile or mint tea. \\
    As there are equal amounts of chamomile tea and mint tea, there is a total of 12 cups / 2 = <<12/2=6>>6 cups of mint tea. \\
    \#\#\#\# 6 \\
    The answer is: 6  \\
    \\
    \textbf{Intermediate Question 3}: How many cups of mint tea are in each row? 
    \textbf{Intermediate Solution 3}:If there are 15 cups of cinnamon tea, then there are a total of 27 - 15 = <<27-15=12>>12 cups of chamomile or mint tea. \\
    As there are equal amounts of chamomile tea and mint tea, there is a total of 12 cups / 2 = <<12/2=6>>6 cups of mint tea. \\
    Dividing these into rows shows that each row holds 6 / 3 = <<6/3=2>>2 cups of mint tea. \\
    \#\#\#\# 2 \\
    The answer is: 2 \\
    \\
    Please strictly follow the format I give you. 
    \\
\bottomrule
\caption{The prompt input to GPT-3.5 for question decomposition on GSM8K dataset} \label{tab:prompt.GSM}
\end{longtable}

\newpage
\begin{longtable}{p{15.7cm}}
\toprule
    Let’s generate the subquestions(subinstructions) and subsolutions to obtain the final answer for this math problem. Use exactly one operation per step. Mathematical expression should be in latex format (e.g. $\binom{5}{2}5^3$). Put your final answer in a box (e.g. $\boxed{\frac{625}{648}}$).
    \\
    \\
    Example 1:
    \\
    \\
    \textbf{Original Question}: Find the value of $n$ that satisfies $2(n+1)!+6n!=3(n+1)!$, where $n! = n \cdot (n-1) \cdot (n-2) \cdots 2 \cdot 1$. 
    \\
    \textbf{Groundtruth answer}: 5 
    \\
    \textbf{Subquestion 1}: Move all terms to the right side. \\
    \textbf{Subsolution 1}: Moving all terms to the right side: $$0=3(n+1)!-2(n+1)!-6n!$$ $$\boxed{0=(n+1)!-6n!}$$ 
    \\
    \textbf{Subquestion 2}: Take out a factor of $n!$. 
    \\
    \textbf{Subsolution 2}: $$0=n!(n+1-6)$$ $$0=n!(n-5)$$ 
    \\
    \textbf{Subquestion 3}: Divide out $n!$. \\
    \textbf{Subsolution 3}: We know that $n!\neq0$, so we can divide out $n!$ and solve for $n$: $$0=n-5$$ $$n=\boxed{5}$$ 
    \\
    Example 2:
    \\
    \\
    \textbf{Original Question}: Steve has one quarter, two nickels and three pennies. Assuming no items are free, for how many different-priced items could Steve individually pay for with exact change? 
    \\
    \textbf{Groundtruth answer}: 23 
    \\
    \textbf{Subquestion 1}: How many possibilities does the quaters contribute? \\
    \textbf{Subsolution 1}: Steve can use no quarters or one quarter, for $\boxed{2}$ possibilities. 
    \\
    \textbf{Subquestion 2}: How many possibilities do the nickels provide? 
    \\
    \textbf{Subsolution 2}: Steve can use 0, 1, or 2 nickels, for $\boxed{3}$ possibilities. \\
    \\
    \textbf{Subquestion 3}: How many possibilities will the pennies provide? \\
    \textbf{Subsolution 3}: Steve can use 0, 1, 2, or 3 pennies, for $\boxed{4}$ possibilities. \\
    \textbf{Subquestion 4}: How many possibilies in total? \\
    \textbf{Subsolution 4}: That gives $2 \cdot 3 \cdot 4 = 24$ possible combinations.  But we must remove the combination where Steve does not use any coins, leaving us with $24 - 1 = \boxed{23}.$
    \\
    \\
    \textbf{Original Question}: \\
    \textbf{Groundtruth answer}: \\
\bottomrule
\caption{The prompt input to GPT-3.5 for question decomposition on MATH dataset} \label{tab:prompt.GSM}
\end{longtable}

\newpage
\begin{longtable}{p{15.7cm}}
\toprule
Here are several facts that will help answer a question. Please organize an inference with the given facts to answer the question, with an explicit answer of either True or False at the end. Then break the question down into several intermediate questions that ask the stage of each intermediate step of your inference. For example: \\
\\
\textbf{Original Question}: Do the anchors on Rede Globo speak Chinese? \\
\\
\textbf{Facts}: \\
1. Rede Globo is a Brazilian television network. \\
2. The official language of Brazil is Portuguese. \\
\\
\textbf{Inference Solution}: No. Rede Globo is a Brazilian television network, and Brazil's official language is Portuguese. Thus anchors on Rede Globo do not speak Chinese. The answer is: False. \\
\\
\textbf{Break Down}: \\
\textbf{Intermediate Question 1}: What country broadcasts Rede Globo? \\
\textbf{Intermediate Solution 1}: Rede Globo is a Brazilian television network. \\
The answer is: Brazil. \\
\\
\textbf{Intermediate Question 2}: What is the official language of Brazil? \\
\textbf{Intermediate Solution 2}: The official language of Brazil is Portuguese. \\
The answer is: Portuguese. \\
\\
\textbf{Intermediate Question 3}: Is Portuguese Chinese? \\
\textbf{Intermediate Solution 3}: The Portuguese is not Chinese. \\
The answer is: False. \\
\\
Please strictly follow the format I give you. Generate the Inference Solution first, and then break down the question into several Intermediate Question and Intermediate Solution. \\

\bottomrule
\caption{The prompt input to GPT-3.5 for question decomposition on StrategyQA dataset} \label{tab:prompt.QA}
\end{longtable}



\newpage
\subsection{AutoPRM Model Prompt}
\label{sec:appendix_QD}

\subsubsection{Question Decomposition (QD) Model Prompt}

\begin{longtable}{p{15.7cm}}
\toprule
    Let's break down this question into a chain of subquestions. \\
    \textbf{Context}: Tobias is buying a new pair of shoes that costs \$95. He has been saving up his money each month for the past three months. He gets a \$5 allowance a month. He also mows lawns and shovels driveways. He charges \$15 to mow a lawn and \$7 to shovel. After buying the shoes, he has \$15 in change. \\
    \textbf{Question}: If he mows 4 lawns, how many driveways did he shovel?
    \\
    \\
    \textbf{Chain of subquestions}: How much money did John save up in total? -> How much money did Tim save from his allowance? -> How much money did John earn from mowing lawns? -> How much money did John earn from shoveling driveways? -> How many driveways did he shovel?
    \\
\bottomrule
\caption{The prompt input to AutoPRM for question decomposition (QD)} \label{tab:prompt_QD}
\end{longtable}

\subsubsection{Question Answering (QA) Model Prompt}
\begin{longtable}{p{15.7cm}}
\toprule
    Below is a math question. Write a solution that answers to the question. The solution may not use all the conditions provided in the question.
    \\
    \\
    \textbf{Question}: \\
    \textbf{Solution} \\
\bottomrule
\caption{The prompt input to AutoPRM for question answering (QA)} \label{tab:prompt_QA}
\end{longtable}

\subsection{Process-supervised Data Annotations}\label{sec:appendix_annotation}

In this section, we detail the data annotation procedure to train Process-supervised Reward Model (PRM) baselines~\cite{lightman2023lets, uesato2022solving} to compare with our proposed model. As outlined in Section \ref{sec:experiment_setup}, the PRM is trained using step-wise labels to assess the correctness of each step. We gather this data by having human annotators review the original question and standard solution from the arithmetical and commonsense reasoning dataset (GSM8K~\cite{cobbe2021training}, MATH~\cite{hendrycks2021measuring}, and StrategyQA~\cite{geva2021did}),  as well as the solution generated by the model. Specifically, annotators are asked to identify the first step in the model solution that contains a significant error, if any. A significant error, as defined in existing works~\cite{uesato2022solving}, is a step where the reasoning is either incorrect or makes it impossible to reach the correct solution anymore without revising that step. Based on these assessments, each step receives a binary label: steps preceding the first significant error are marked as 'correct', while subsequent steps are labeled 'incorrect'.

\subsection{Data Preparation for 
 Interpreting Decomposition Granularity}
\label{sec:appendix_granularity}

To prepare data to train AutoPRM to interprete with the decomposition granularity parameter, we follow~\cite{han2023lm} and assign $\epsilon=1$ to all the fully-decomposed subquestion-subsolution pairs and $\epsilon=0$ to the one-shot CoT prompt of each problem $p_i$. Specifically, for an intermediate $\epsilon$, we select a subset of $\tilde{n}_i$ subquestions from
the fully-decomposed pairs that significantly contribute to the final answer via a heuristic, 
and assign $\epsilon=\tilde{n}_i/n_i$. 

The heuristic determines if a subsolution in a reasoning process is important by checking if it introduces a new condition 
 or calculation into the reference context. Specifically, we adopt a simple token-matching method via regex to sequentially check for new conditions. For example, for GSM8K we extract two types of tokens: entities and numbers. Sequentially, we check if each subsolution in the decomposed subquestion-solution pairs set introduces a new entity (implying new condition) or new number (implying calculation). If yes, we consider this subsolution as contributive to the final-answer.
Finally, this $\epsilon$ is integrated into the QD prompt and optimized using 
Eqn.~(\ref{eq:sft_loss}).

\end{document}

%% file: content/introduction.tex
\section{Introduction}
The landscape of natural language processing has been profoundly reshaped by
the evolution of large language models (LLMs), which have demonstrated remarkable 
capabilities in a variety of complex 
tasks~\cite{brown2020language, chen2021evaluating, yuan2023scaling}.
Among these, multi-step reasoning has emerged as a particularly challenging 
area and has drawn significant research 
attention~\cite{bhattacharya2017survey, hoffmann2022training, bubeck2023sparks}.
To enhance complex reasoning capabilities for LLMs, recent prompting-based approaches, including chain-of-thought 
(CoT)~\cite{kojima2022large, wei2022chain} and
self-evaluation decoding~\cite{wang2022self, yao2023tree, xie2023decomposition},
have proven to be successful. 
However, while being effective on large-sized models (\emph{e.g.}, \texttt{GPT-4}~\cite{openai2023gpt4}, \texttt{PaLM-2}~\cite{anil2023palm}), they are less effective 
for smaller-sized non-finetuned models (\textit{e.g.}, \texttt{GPT-3}~\cite{brown2020language}, \texttt{LLaMA-2-7B}~\cite{touvron2023llama}) which are poor reasoners by nature~\cite{stolfo2022causal}.

\begin{figure}[t]
    \centering
    \includegraphics[width=0.48\textwidth]{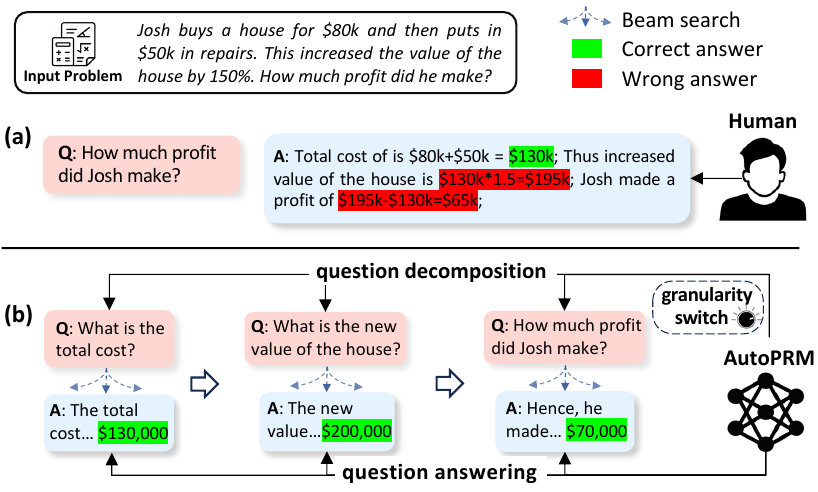}
    \caption{The decoding pipeline of our proposed AutoPRM, which consists of a unified question decomposition (QD) and question answering (QA) model. First, QD breaks down the problem into a series of sub-questions according to a user-specified granularity. Then, the RL-optimized QA model solves them sequentially via FiM-decoding, which consistently guides QA toward the solution of the primary problem. 
    }
    \label{fig:decoding}
    \vspace{-0.15in}
\end{figure}

On the other hand, fine-tuning methods are also known to be effective for enhancing complex
reasoning capabilities, especially for smaller-sized 
models~\cite{uesato2022solving, luo2023wizardmath, shridhar2023distilling}.
Therein, procedural supervision-based fine-tuning~\cite{wu2023fine, lightman2023lets} has proved to be particularly 
effective~\cite{uesato2022solving}, which emulates human problem-solving 
process and provides step-by-step feedback, as opposed to outcome-based supervision which simply optimizes for the final-answers~\cite{shridhar2023distilling}.
Despite its advancements, the reliance on step-wise human annotations for these
process-supervised reward models (PRM) presents a significant bottleneck.
More specifically, such annotation is not only resource-intensive in terms of both 
time and domain expertise~\cite{lightman2023lets}, but also introduces human bias due to 
subjective judgements, which could potentially undermine the fine-tuning 
performance~\cite{casper2023open, lightman2023lets}.

These challenges highlight a critical need for a more efficient and scalable approach to fine-tune smaller-sized LLMs for complex reasoning tasks.
%
To address these challenges, we propose a novel self-supervised 
procedural reward model termed \textbf{AutoPRM} to boost the efficiency and accuracy in fine-tuning and inference for long-chained reasoning problems.
Concretely, AutoPRM first breaks down the problem into a sequence of
sub-questions with a trained question-decomposition (QD) model, then solves each subquestion with a Reinforcement Learning (RL)-optimized question-answering (QA) model~\cite{silver2017mastering}.  

Inspired by human cognitive process where questioning and answering reciprocally enhances each other~\cite{xu2023effectiveness}, we train one unified model to handle both QD and QA.\footnote{Note that while we refer to QD and QA separately, they refer to the dual functions of a unified model in the remaining of the paper.} Notably, instead of training a PRM that requires extensive manual annotations, AutoPRM adopts a more natural approach by directly training an intermediate outcome verifier to optimize the subquestion solver.

%


The main contributions of this paper are: (1) AutoPRM, a novel fine-tuning framework that enhances 
LLMs reasoning abilities.
This framework reduces the need for extensive human annotations by employing automatic question 
decomposition and an RL-optimized subquestion solver.
%
%
(2) Through extensive experiments conducted on two arithmetic reasoning datasets, 
GSM8K~\cite{cobbe2021training}, MATH~\cite{hendrycksmath2021}, and one commonsense reasoning 
dataset StrategyQA~\cite{geva2021did}, we demonstrate AutoPRM's effectiveness on improving LLMs generic
multi-step reasoning capabilities. 
%

%



%% file: content/method.tex
\section{AutoPRM}\label{sec:method}
In this section, we detail our proposed framework \textbf{AutoPRM}. Our key insight is that automating step-wise question decomposition provides a natural perspective to reduce problem dimensions,
through which model inference and optimization can be more precise and efficient. With decomposition, the fine-grained feedback can be obtained with a reliable step-wise verifier trained to predict intermediate results, which lead to a more powerful and bias-free reasoning model.

\noindent \textbf{Preliminaries and Notations.} Our problem formulation involves a dataset 
$\mathcal{D} = \{(p_i, a_i)\}_{i=1}^N$, where each problem $p_i$ is 
associated with a final answer $a_i$ that can be reached through reasoning. 
AutoPRM breaks the problem into multiple steps and models the reasoning 
process as a Markov Decision Process (MDP) 
$\langle \mathcal{S}, \mathcal{A}, \mathcal{Q}, \mathcal{R}, P, \gamma \rangle$~\cite{ramamurthy2022reinforcement}, where each MDP 
episode starts with a sampled problem input $p_i$ and ends either when a 
final answer is generated or the model abstains. 
AutoPRM manages each individual reasoning step as a subquestion-solution pair $\{(q_t,s_t),q_t \in \mathcal{Q}, s_t \in \mathcal{S}$\}, where $\mathcal{S}$ is the state (subsolution) space and 
$\mathcal{Q}$ is the subquestion space.\footnote{Note that while state $s_t$ should strictly refer to the cumulative subsolutions $\sum_0^t{s_i}$  (and action $a_t \in \mathcal{A}$ denotes the $j^\text{th}$ subsolution), we represent $s_t$ as the $t^\text{th}$ subquestion for the simplification of notation.}
The transition function appends 
a subsolution $s_t$ to the end of the cumulative state 
$(p_i, s_0, s_1, \ldots, s_{t-1})$ at each step. 
A reward function (verifier)
$\mathcal{R}: \mathcal{Q} \times \mathcal{S} \rightarrow \mathbb{R}$ can be 
either outcome-based~\cite{cobbe2021training}, which provides a sparse 
feedback at the end of generation, or process-based~\cite{uesato2022solving}, 
which evaluates each step individually and assigns a fine-grained score for each intermediate step $(q_t, s_t)$.
%

\subsection{Collecting Subquestion-Subsolution Pairs}\label{subsec:automated_dqc}

In this subsection, we illustrate the procedures to prepare the subquestion-subsolution dataset
$\mathcal{D}_{\text{sub}} = \{\{q_{i, t}, s_{i, t}\}_{t=1}^{n_i}\}_{i=1}^N$.
 $\mathcal{D}_{\text{sub}}$ will be used to train 
the QD and QA models, which are the core parts of the proposed AutoPRM.

Instead of manually providing subquestions as seen in prior
works~\cite{xie2023decomposition}, AutoPRM adopts an efficient and unified framework to 
collect subquestions by training an auxiliary subquestion collection (SQC) Model. 
The core idea behind SQC is based on the assumption that each sentence in the groundtruth 
solution represents a valid step that progressively leads to the final
solution~\cite{lightman2023lets}. 
In practice, we treat each sentence in the groundtruth solution as a subsolution, and take these
subsolutions as inputs to SQC to generate the corresponding subquestions.
%
%
 
To train this SQC model, we initially select a small subset of the original dataset 
$\mathcal{D}$ and prompt GPT-3.5 (prompts are detailed in~\ref{sec:appendix_subQG}) to find appropriate subquestions that 
would reasonably induce these subsolutions. 
Next, we fine-tune an open-source Language Model (LM) (e.g. LLaMA-2~\cite{touvron2023llama}) 
on this training set to obtain the SQC model.
%
We employ the SQC model to break each question-solution pair in the original dataset 
$\mathcal{D}$ into corresponding subquestions and subsolutions, yielding a new dataset 
$\mathcal{D}_{\text{sub}}$.

\begin{figure*}[ht]
    \centering
    \includegraphics[width=\textwidth]{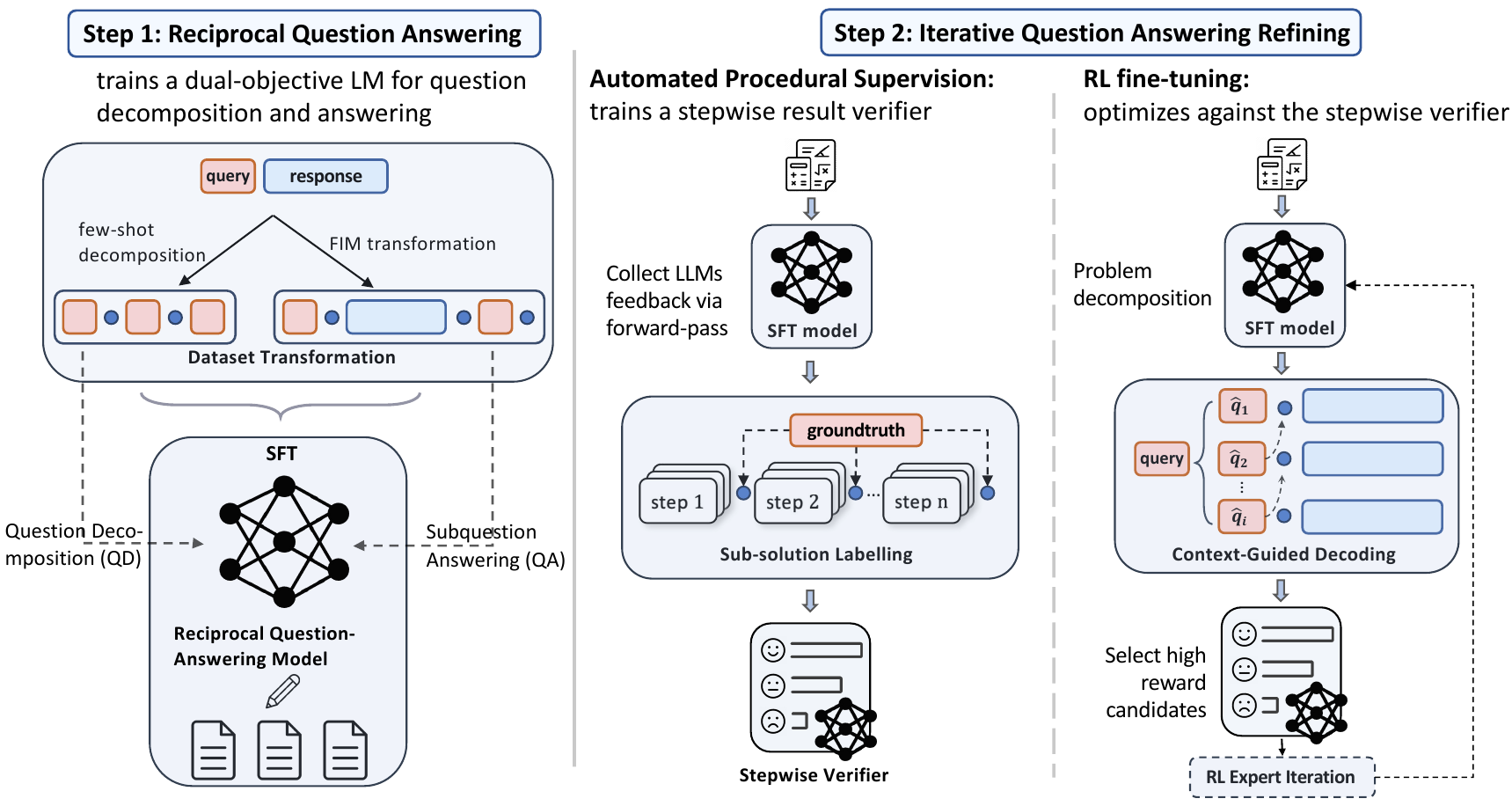}
    \caption{A diagram illustrating the 3 steps of AutoPRM: (1) supervised ﬁne-tuning (SFT) on a merged dataset of question decomposition dataset $\mathcal{D}_{QG}$ and the FIM-transformed question answering dataset $\mathcal{D}_{QA}$; (2) stepwise result verifier trained on the LLMs generated solutions of $\mathcal{D}_{QA}$; (3) RL fine-tuning against the stepwise verifier. The base model first decomposes the question into several intermediate subquestions and solve them sequentially via LGD. Then the candidates with high reward are selected to fine-tune the policy via expert iteration.}
    \label{fig:pipeline}
    \vspace{-0.1in}
\end{figure*}

\subsection{Reciprocal Question Answering}\label{subsec:learning_qd}
The core idea of our proposed AutoPRM is to fine-tune LLMs via automatically generated 
intermediate solutions.
This process is consist of two parts, which are question decomposition (QD) and question 
answering (QA).
More specifically, QD divides each question into a sequence of subquestions, while QA answers
these subquestions and generates corresponding subsolutions.

Both QD and QA naturally connect and influence each other. 
Therefore, to leverage the interconnections between them and better refine the overall 
effectiveness of AutoPRM, we propose \textit{Reciprocal Question Answering (RQA)}.
In fact, RQA is inspired and theoretically grounded by the cognitive learning theory stating 
that a mutual enhancement does exist between learning to ask relevant questions and solving 
problems, or in other words, improved problem-solving skills lead to more precise 
questions, and vice versa~\cite{xu2023effectiveness}.
%

%
%
Different from SQC which relies on the ground truth 
solutions to obtain the subquestions, our QD is designed to automatically break down any 
arbitrary question without access to the ground truths.
Precisely, the QD model takes each problem $p_i$ as input and 
generates a set of decomposed subquestions 
${(q_t, s_t)}_{t=1}^{n_i}$. 
However, exclusively training this QD model may result in overfitting, potentially diverting the model 
from its primary purpose of facilitating multi-step reasoning.
To tackle this challenge, we encompass the original QD objective with the additional QA 
objective, which aims to infer subsolutions based on the corresponding subquestions and their 
surrounding context. 

In practice, we adopt two separate prompting mechanisms specified for QD and QA 
(prompt detailed in~\ref{sec:appendix_QD}). 
For QD, during the training phase, we concatenate each subquestion with a special split token 
to form an ordered sequence of subquestions.
And in inference, we use the same prompt but parse the output directly into a set of 
subquestions.
%
In terms of QA, we propose \textit{Context-Guided Decoding (CGD)} that is similar to the
Fill-In-the-Middle (FIM) Decoding~\cite{li2023starcoder, liang2023code}, where each 
subquestion solver is guided by appending the subsequent subquestion to the beginning of the 
subsolutions that have being inferred.
The detailed formulation of CGD is showed in Eqn~(\ref{eq:cgd}), with $\circ$ denoting the
concatenation.
%
\begin{equation}
    \langle \texttt{PRE} \rangle \circ q_{t} \circ \langle \texttt{SUF} \rangle \circ q_{t+1} \circ \langle \texttt{MID} \rangle \circ [s_0, \ldots, s_{t-1}]
\label{eq:cgd}
\end{equation}
Through CGD, we can train the QA model to derive holistically rational subsolutions that respond 
well to $q_t$ while making good progress towards the final solution, instead of focusing only on the partial context associated with
$q_t$ and deviating from the original task. 

Based on these two prompting mechanisms, we construct an auto-regressive language modeling loss 
$\mathcal{L}(\mathcal{D}_{\text{sub}})$:
\begin{equation}
\begin{aligned}
\mathcal{L}(\mathcal{D}_{\text{sub}}) &= \sum_{i=1}^{N} \Biggl( \underbrace{- \sum_{t=1}^{n_i} \log P(q_t | q_{<t}, p_i)}_{\text{QD Loss}} \\
&\quad \underbrace{- \sum_{t=0}^{n_i} \log P(s_t | s_{<t}, q_t, q_{t+1})}_{\text{QA Loss}} \Biggr)
\end{aligned}
\label{eq:sft_loss}
\end{equation}
where $P(q_t | q_{<t}, p_i)$ is the probability of generating subquestion $q_t$ conditioned on 
all the previous subquestions $q_{<t}$ and input problem $p_i$, aligning with the QD objective.
And $P(s_t | s_{<t}, q_t, q_{t+1})$ is the probability of the model generating the subsolution $s_t$ 
conditioned on all the preceding subsolutions $s_{<t}$, current subquestion $q_t$ and subsequent 
subquestion $q_{t+1}$, aligning with the QA objective.

Empirically, we observe that some of the decomposed subquestion-subsolution pairs obtained in 
\S\ref{subsec:automated_dqc} are redundant and do not significantly contribute to the 
original final solutions. 
To this end, we propose a user-defined parameter $\epsilon \in [0, 1]$ to manage 
the granularity of the decompositions.
%
Switching between decomposition granularity is tied to the speciﬁc choice of words within the problem context, which is equivalent to a linear transform in the embedding space~\cite{han2023lm}. 
%

When $\epsilon=1$, the original question-solution pairs are fully-decomposed into the
subquestion-subsolution pairs, while decomposition with $\epsilon=0$ being practically the same as 
the one-shot CoT results.
For an intermediate $\epsilon$, we select a subset of $\tilde{n}_i$ subquestions from
the fully-decomposed pairs that significantly contribute to the final answer via a heuristic, 
and assign $\epsilon=\tilde{n}_i/n_i$. 
Finally, this $\epsilon$ is integrated into the QD prompt and optimized using 
Eqn.~(\ref{eq:sft_loss}).
We refer more details to~\ref{sec:appendix_granularity}.

\subsection{Question Answering Refining}
\subsubsection{Automated Procedural Supervision}
\label{subsec:rl_finetuning}
%
While the reciprocal QA model trained in Section \ref{subsec:learning_qd} is sufficient to handle complex reasoning problems, they suffer from partial context and a loss of holistic view to obatin the final answer, due to decomposition. Thus we aim to further enhance the QA model  with automated procedural supervision via RL.
Specifically, this is achieved by first letting the QD model decompose problem $p_i$ into 
intermediate subquestions, and then acquiring feedback $r_t$ for each subsolution $s_t$ via a 
step-wise binary verifier.
%
%
As shown in Figure~\ref{fig:pipeline}, we implement a step-wise verifier (reward model, RM) 
as a language model to predict a binary label as either a `correct' or `incorrect' token after 
each step. 
To train this verifier, we first obtain a sample set of subquestion-subsolution pairs, as shown 
in the middle step of Figure~\ref{fig:pipeline}. 
Then the intermediate result of each subsolution is compared with the groundtruth 
and assigned values based on:
\begin{equation}
\label{eq:QA_indicator}
    I(\text{QA}(s_{t,i}), a_{t,i}) = 
    \begin{cases}
        1 & \text{if } \text{QA}(s_{t,i}) = a_{t,i}, \\
        0 & \text{otherwise},
    \end{cases}
\end{equation}
where $a_{t,i}$ is the intermediate groundtruth answer for the $t^{\text{th}}$ subquestion of 
problem $p_i$.
Notice that the policy that maximizes the score of intermediate steps also maximizes the 
RM-estimated probability of eventually reaching the correct ﬁnal answer. 
Finally, the output label from Eqn.~(\ref{eq:QA_indicator}) 
is appended after each subsolution and trained via the QA loss as defined in 
Eqn.~(\ref{eq:sft_loss}).

\subsubsection{RL Fine-tuning}
\label{subsec:rl_finetuning}
After obtaining the step-wise verifier, the last step of our pipeline, as showed in 
Figure~\ref{fig:pipeline}, is to apply RL via expert iteration~\cite{silver2017mastering} to 
further fine-tune the SFT models. 
Different from policy gradient methods, expert iteration alternates between policy improvement 
and policy distillation. 
In policy improvement, QA model produces $k$ candidates for each problem $p_i$ in the dataset 
$\mathcal{D}_{\text{sub}}$ via a decoding method. 
Then in policy distillation, we select the candidates with the highest scores based on the 
verifier and perform supervised-learning to improve the policy.
%

%

When the training converges, we adopt Reward-Reranking (RR) decoding which selects the candidate 
subsolutions via RM-weighted probability~\cite{uesato2022solving, xie2023decomposition} as the new 
score. 
Mathematically, we have:
%
\begin{equation}
s_t = \arg\max_{s_t} P_{\mathcal{M}}(s_t|s_{<t},q_t,q_{t+1}) \cdot \mathcal{R}(s_t)
\label{eq:score}
\end{equation}
where $\mathcal{R}(s_t)$ is the probability for predicting "correct" by reward model $\mathcal{R}$. 
The insight here is that a correct reasoning step should be confirmed by both the inference model and the verifier.

In the decoding process, RR is coupled with beam search (BS), a step-wise tree-based algorithm, to 
select the most probable sequence of words or tokens.
%
Specifically in AutoPRM, as QA generates $m$ candidate steps for each decomposed subquestion $q$, 
the top $k$ candidates are selected according to their decoding scores as showed in 
Eqn.~(\ref{eq:score}).
This process is repeated until the model outputs the ﬁnal answer or abstains from answering. 
%
The abstaining condition triggers when the scores of all $mk$ candidates drop below the threshold 
$\tau$ at any step~\cite{geifman2017selective}. 
The complete AutoPRM decoding procedure for multi-step reasoning is illustrated in 
in Algorithm~\ref{algo:decoding}. Through these strategies, we can eventually build a reliable and efficient QA model.
\begin{algorithm}[t!]
\caption{AutoPRM Decoding Procedure}
\begin{algorithmic}[1] 
  \Require{Fine-tuned model $\mathcal{M}$, problem $p$, granularity parameter $\epsilon$, abstaining threshold $\tau$.}
  
\State The model $\mathcal{M}$ decomposes the problem $p$ into multiple sub-questions $\{q_1,q_2,...,q_N\}.$
\For{$t=1,...,N$}
    \State Take each one of the $k$ generated solution $s_{<t}$ as prefix, append $q_{t+1}$ to the end of $s_{t}$ via context-guided decoding
    \State Sample $m$ candidates of $s_{t}$
    \If{the score of all $mk$ candidates are less than $\tau$}
        \State Abstain.
    \Else
    \State Choose top $k$ candidates of $s_t$ from all $mk$ samples according to the score \eqref{eq:score}.
    \EndIf
\EndFor
\end{algorithmic}
\label{algo:decoding}
\end{algorithm}

%% file: content/experiment.tex
\section{Experiments}\label{sec:experiment}

In this section, we assess AutoPRM, our proposed framework, by exploring three
key questions: 
(1) to what extent does AutoPRM enhance the reasoning capabilities of LLMs?
(2) how do individual sub-modules contribute to AutoPRM's overall performance 
improvement? 
and (3) what are the limitations and opportunities in enhancing multi-step reasoning for smaller-scaled models?

\subsection{Experimental Setups}
\label{sec:experiment_setup}
\paragraph{Datasets.}

We assess AutoPRM on two arithmetic reasoning datasets, namely GSM8K~\cite{cobbe2021training}, 
MATH~\cite{hendrycks2021measuring}, and one additional commonsense reasoning dataset,
StrategyQA~\cite{geva2021did} in our experiments.

GSM8K~\cite{cobbe2021training} features 8.5k grade-school-level math problems, which is 
ideal for assessing AutoPRM's basic arithmetic reasoning skills.
And MATH~\cite{hendrycks2021measuring}, on the other hand, contains 12.5k more complex 
problems that covers a wider range of mathematical topics, further challenging
LLMs' ability to conduct complex mathematical reasoning.
%

On the other hand, StrategyQA~\cite{geva2021did} serves to assess AutoPRM's capability for 
commonsense reasoning, which involves understanding implicit assumptions and making inferences based on contexual information and internal model knowledge to answer questions that are not strictly mathematical but rely heavily on logical 
thinking.
We follow existing work~\cite{shridhar2023distilling} for data split and pre-processing.



\paragraph{Baselines.}

We compare AutoPRM with a wide range of SOTA models. WizardMath~\cite{luo2023wizardmath} and MetaMath~\cite{yu2023metamath} are two SOTA models that enhance mathematical reasoning with external data augmentation. Distilling-LM~\cite{shridhar2023distilling} is a decomposition-based reasoning framework that adopts two separate models for QD and QA, and then distills (SFT) reasoning capabilities from GPT-3.5. 

As for reward-based (verifier-based) models, 
we consider ORM-RL~\cite{cobbe2021training} and PRM-RL~\cite{lightman2023lets} as our baselines, and follow the exact training procedures outlined in their papers.
Except for LLaMA-2-70B model that adopts a few-shot decoding without fine-tuning, all other approaches have been fine-tuned based on LLaMA-2-7B~\cite{touvron2023llama}\footnote{Since Distilling-LM did not publish its results on LLaMA-2 nor its
training data, we report the result using our dataset.}.

%

%

\paragraph{Experimental Settings.} 
To fairly compare with PRM baselines, we follow \citet{uesato2022solving} and annotate a 
comparative amount of procedural feedback data equivalent to the subquestion-subsolution 
pairs used to train our step-wise verifier in AutoPRM. Please refer to Appendix~\ref{sec:appendix_annotation} for detailed annotation 
procedures.

During decoding, WizardMath~\cite{luo2023wizardmath} and MetaMath~\cite{yu2023metamath} adopt one-shot greedy decoding, while other models adopt a beam search of size eight. We consistently set granularity $\epsilon=0.8$ across all tests. In addition, the iterative process of RL through expert iteration was conducted over five epochs, with the
best model being selected based on its performance in final-answer error on the 
validation set.
All model training was conducted using Huggingface~\cite{wolf2020transformers}. The detailed hyperparameters are reported in \ref{sec:appendix_hyperparameters}.

\subsection{Results on Arithmetic Reasoning}
\label{sec:main}

\begin{table}
\centering
\tabcolsep=0.15cm
\begin{tabular}{lll}
\hline
\textbf{Model} & \textbf{GSM8K} & \textbf{MATH} \\
\hline
LLaMA-2 (70B) & 56.8 & 13.5 \\
LLaMA-2 (7B) & 41.6 & 4.7 \\
WizardMath* & 54.9 & 10.7\\
MetaMath* & 66.4 & 19.4\\
Distilling-LM & 51.8 & 10.2\\
ORM-RL & 52.9 & 6.9\\
PRM-RL& 56.1 & 10.5\\
\hline
\textbf{AutoPRM} & \textbf{59.3 (+3.2)} & \textbf{13.2 (+2.7)} \\
\textbf{AutoPRM}* & \textbf{70.8 (+4.4)} & \textbf{23.6 (+4.2)} \\ 
\hline
\end{tabular}
\caption{
Comparison on GSM8K and MATH dataset. All models are fine-tuned based on LLaMA-2 7B by default.
* indicates SFT on external data.}
\label{tab:gsm8k_result}
\vspace{-0.3in}
\end{table}


For arithmetic reasoning tasks, we have trained two sets of models: one is based on the groundtruth training dataset provided in GSM8K~\cite{cobbe2021training} and MATH~\cite{hendrycks2021measuring}, and the other one is augmented using MetaMath~\cite{yu2023metamath}.
The results are reported in Table~\ref{tab:gsm8k_result}.

Results indicate that AutoPRM achieves the best performance compared with other models. Specifically, AutoPRM reaches 59.3\% on GSM8K and 13.2\% on MATH with beam search, which outperforms PRM-RL by 3.2\% and 2.7\% respectively. 
Additionally, when applying our proposed pipeline to the MetaMath dataset for data augmentation
and fine-tuning the model~\footnote{\url{https://huggingface.co/meta-math/MetaMath-7B-V1.0}},
AutoPRM's effectiveness is further improved, reaching 70.8\% (+4.4\%) on GSM8K 
and 23.6\% (+4.2\%) on MATH. 
These results highlight AutoPRM's substantial performance improvements and its versatility 
when combining with other methods. 
%

%



%

\subsection{Results on Commonsense Reasoning}
For the commonsense reasoning task tested with StrategyQA~\cite{geva2021did}, we consider the fine-tuned model on LLaMA-2-7B as our baseline. For ORM-RL, we directly train an outcome verifier based on the final binary prediction. And for PRM-RL, we follow \citet{uesato2022solving} and annotate step-by-step to train a procedural supervised reward. The results are reported in Table~\ref{tab:limitation_strategyQA}.

The final-answer accuracy is verified with the groundtruth binary label. And to supervise 
step-wise (trace) result, we transform all open-ended subquestions to be close-ended, which can be answered with either "yes" or "no". Results confirm that procedural-supervised methods outperform outcome-based methods, and AutoPRM further achieves an accuracy gain of 1.2\% on final-answer and 1.4\% on intermediate result thanks to our precise and unbiased feedback. While the positive result indicates the effectiveness of our proposed method, the performance gain is less significant comparing to the large gains we observe in the arithmetic reasoning tasks. 

We suspect that such result could be due to the model knowledge 
gap~\cite{petroni2019language}, as StrategyQA highly depends on factual truthfulness, 
which cannot be further enhanced by simply improving the reasoning framework.
Therefore, we believe increasing model scale is the key factor to further improve accuracy 
on such knowledge-dependent reasoning tasks~\cite{anil2023palm}.

\label{sec:QA_result}

\begin{table}
\centering
\tabcolsep=0.03cm
\begin{tabular}{lcc}
\hline
\textbf{Approach} & \textbf{Final-Answer} & \textbf{Trace Result} \\
\hline
SFT & 56.8 & 61.0 \\
SFT+ORM-RL & 58.2 & 60.5 \\
SFT+PRM-RL & 65.1 & 66.0 \\
\textbf{SFT-AutoPRM-RL} & \textbf{66.3 (+1.2)} & \textbf{67.4 (+1.4)} \\

\hline
\end{tabular}%
\caption{
The accuracy of final-answer and stepwise (trace) result on StrategyQA dataset. The result indicates that while AutoPRM still improve upon ORM and PRM-based methods, its improvement is not as significant as in arithmetic reasoning tasks, probably due to model scale.}
\label{tab:limitation_strategyQA}
\vspace{-0.15in}
\end{table}

\subsection{Analysis}
In this section, we investigate the contributions of each sub-module to the improvement of AutoPRM, including decoding methods and controllable question decomposition. Additionally, we analyze the trace error and examine the correlation between AutoPRM's performance and problem length.
%

\noindent \textbf{Decoding Methods.} 
\begin{table}
\centering
\tabcolsep=0.2cm
\begin{tabular}{lcc}
\hline
\textbf{Approach} & \textbf{GSM8K} & \textbf{MATH}\\
\hline
SFT & 41.6 & 4.7\\
ORM-RL+Greedy & 45.4 & 5.1\\
ORM-RL+SC & 51.9 & 6.9\\
PRM-RL+Greedy& 51.3 & 7.4\\
PRM-RL+BS & 56.1 & 10.5\\
\textbf{AutoPRM+Greedy} & \textbf{52.8} & \textbf{10.4}\\
\textbf{AutoPRM+BS} & \textbf{58.2 (+2.1)} & \textbf{11.9 (+1.4)}\\
\textbf{AutoPRM+RR} & \textbf{58.9 (+2.8)} & \textbf{12.7 (+2.2)} \\
\textbf{AutoPRM+GD} & \textbf{59.3 (+3.2)} & \textbf{13.2 (+2.7)}\\

\hline
\end{tabular}
\caption{
Comparison of testing accuracy on arithmetic datasets w.r.t. reward models and decoding strategies. All approaches is fine-tuned on LLaMA-2-7B model and use naive greedy decoding. RR refers to Reward Ranking. (+Number) indicates the improvement performance compared to the best baseline PRM-RL+BS.
}
\label{tab:RL_model}
\vspace{-0.15in}
\end{table}
First, we compare AutoPRM's performance against various reward 
models and decoding strategies. 
To start, we analyze the performance of ORM, PRM and AutoPRM coupled with different decoding strategies. From the results in Table \ref{tab:RL_model}, we can observe that:

Since ORM-RL is an outcome-supervised method, we apply self-consistency and use the majority answer as the final-answer. All other methods apply a beam search process to yield the final-answer. Our beam search baseline selects candidates by maximizing the RM score at each step, 
while the Reward-Reranking (RR) decoding, as outlined in~\citet{xie2023decomposition}, maximizes the 
RM-reweighted score as showed in Eqn.~(\ref{eq:score}). 

Notice that RR achieves a better performance than beam search since they can select a more 
rational candidate by incorporating signals from both the LM and verifier.
Similar to the RR decoding, our CGD decoding, which guides the individual subsolutions 
towards solving the original problem, can further mitigate the intrinsic issues brought by 
decomposition (\textit{e.g.}, diminished context, topic deviation) and effectively enchance final performance.
%



\label{sec:decoding}

\begin{figure}[t!]
    \centering
    \includegraphics[width=0.45\textwidth]{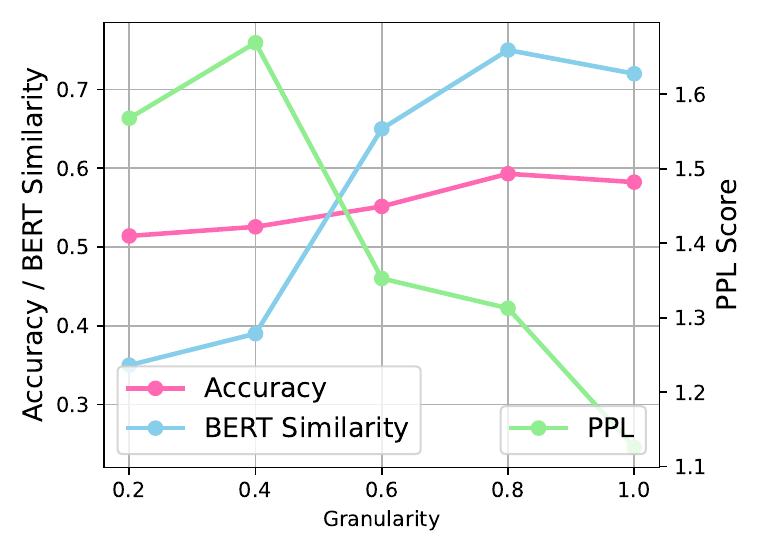}
    \caption{Assessment on GSM8K dataset w.r.t decomposition granularity $\epsilon$. We evaluate the final-answer accuracy, perplexity and BERT similarity (to groundtruth solutions). Accuracy demonstrates that an intermediate granularity level ($\epsilon$=0.8) yields best performance. Perplexity denotes that fine-grained guidance enhances the model's certainty in problem-solving. The increased similarity to the groundtruth solutions imply that AutoPRM effectively decompose questions that align with the human labeller. 
    }
    \label{fig:gsm8k_granularity}
    \vspace{-0.15in}
\end{figure}

\noindent \textbf{Decomposition Granularity.}\label{sec:gll} 
We evaluate AutoPRM performance across varying decomposition granularity on GSM8K. 
As shown in Figure~\ref{fig:gsm8k_granularity}, while we can see that more precise and 
fine-grained decomposition generally leads to better accuracy and more certain factual 
inference~\cite{chuang2023dola}, interestingly, an intermediate level of granularity 
($\epsilon=0.8$) achieves the highest accuracy in final answers. 
Additionally, the increased similarity to the groundtruth solutions indicate that our model 
can effectively break down questions in a manner that aligns with human cognitive process to 
solve multi-step reasoning problems.

\begin{table}
\centering
\begin{tabular}{lcc}
\hline
\textbf{Approach} & \textbf{GSM8K} & \textbf{MATH}\\
\hline

SFT+ORM-RL & 53.4 & 6.5\\
SFT+PRM-RL & 70.5 & 13.1\\
\textbf{AutoPRM-SFT} & 71.2 & 14.5\\
\textbf{AutoPRM-SFT+RL} & \textbf{72.0} & \textbf{14.7}\\
\hline
\end{tabular}
\caption{
Comparison of different fine-tuning methods in solving the decomposed dataset of subquestion-solution pairs. The results indicate that models with process-supervision can also solve the individual subquestions effectively. Conversely, the poor performance of ORM methods in solving the subquestions confirms that direct fine-tuning against ORM can lead to correct final-answer with the incorrect reasoning trace.}
\label{tab:decomposed_questions}
\vspace{-0.15in}
\end{table}

\noindent \textbf{Step-wise (Trace) Error Analysis.}
To evaluate the internal reasoning reliability, we assess the performance on each decomposed 
subquestion-subsolution pair and present the results in Table~\ref{tab:decomposed_questions}. 
%
%
%
Results show that while process-supervised models adeptly solves subquestions, AutoPRM gains 
more from precise, unbiased, and fine-grained feedback. 
Additionally, the fact that ORM-based methods only show slight improvements in original 
question solving (Table~\ref{tab:RL_model}), suggests that they struggle to link correct final 
answers with their reasoning traces, which is also discussed in~\cite{lightman2023lets}.

\begin{figure}[t!]
    \centering
    \includegraphics[width=0.42\textwidth]{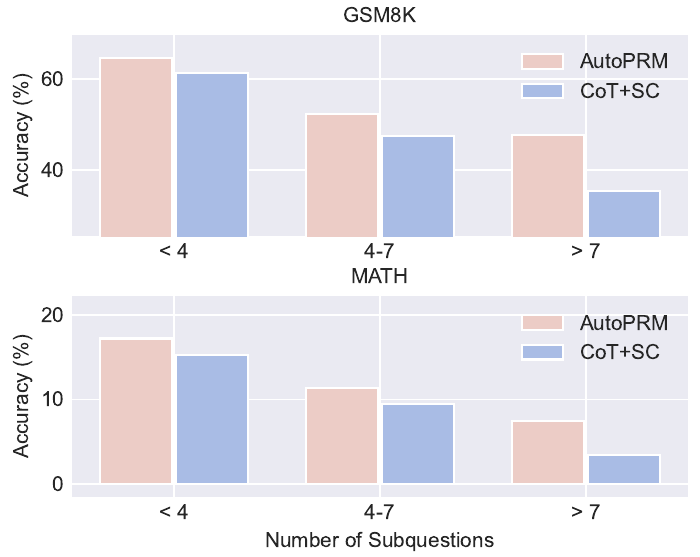}
    \caption{Comparing AutoPRM and CoT+SC decoding on problems of varing complexity and with different number of subquestions. AutoPRM outperforms CoT+SC by a large margin, especially for problems with longer reasoning chains.} 
    \label{fig:comparison_steps}
    \vspace{-0.22in}
\end{figure}

\noindent \textbf{Varying Problem Length.} 
We further access the reasoning performance w.r.t. the length of the
questions, which is a natural reflection of the question complexity and an
approximation of the amount of reasoning needed to derive the final answers. The results are showed in Figure~\ref{fig:comparison_steps}.
%
%
In general, we notice that the performance gain (the absolute accuracy gain over 
self-consistency) increases as the reasoning chain becomes longer, which verifies the 
effectiveness of our method in guiding the reasoning trace to attain the correct final-answer, 
especially for longer-chained problems.

%% file: content/related_work.tex
\section{Related Work}\label{sec:related_work}

LLMs struggle with complex reasoning tasks~\cite{lu2022survey}.
To mitigate this limitation, prompt-based methods including 
Chain-of-Thought (CoT)~\cite{wei2022chain} and its variants such as 
automatic CoT~\cite{zhang2022automatic}, Complex CoT~\cite{fu2022complexity}, 
Tree-of-Thought~\cite{yao2023tree,zhang2024generating}, Graph-of-Thought~\cite{besta2023graph} and 
Exchange-of-Thought~\cite{yin2023exchange} are developed.
Although effective for larger LLMs, the performance of these methods is limited in smaller 
models~\cite{openai2023gpt4, anil2023palm, lewkowycz2022solving}, which leads to the exploration 
of problem decomposition into subquestions and sequential 
handling~\cite{wei2022chain,gao2023pal,chen2022program,zhou2022least}, coupled with step-wise 
feedback using self-verification~\cite{miao2023selfcheck}, 
external LLMs~\cite{miao2023selfcheck, xie2023decomposition}, heuristics~\cite{yao2023tree}, 
and human-annotated rewards~\cite{uesato2022solving, lightman2023lets}.
However, these methods often either depend on external large models, or require intensive 
human effort or specific designs~\cite{lightman2023lets}, which severely limit their 
applicability.

Besides these prompt-based methods, fine-tuning has been another main line of research showing 
promise in enhancing LLMs reasoning capabilities for both large and smaller 
models~\cite{uesato2022solving, luo2023wizardmath, shridhar2023distilling,tian2023fine}, especially 
when pairing with data augmentation techniques such as multi-view question
bootstrapping~\cite{luo2023wizardmath} and instruction evaluation~\cite{yu2023metamath}.
Among numerous fine-tuning approaches, existing studies suggest procedural supervision yields 
better accuracy than outcome-only 
methods~\cite{wu2023fine, lightman2023lets, shridhar2023distilling}.
However, procedural supervision approaches often require extensive, unbiased manual labeling, 
which greatly limits their generalizability~\cite{uesato2022solving}.

Inspired by the step-wise approaches as seen in the prompt-based methods, as well as
procedural supervision used in fine-tuning approaches, in this paper, we introduce a novel
self-supervised fine-tuning approach that significantly enhances LLMs reasoning capabilities while 
not requiring either external large models, or additional human efforts.

%% file: content/conclusion.tex
\section{Conclusions and Future Work}\label{sec:conclusions}

In this paper, we introduce AutoPRM, a novel framework that automates procedural 
supervision for multi-step reasoning in LLMs.
The core of AutoPRM consists of two reciprocal components: a QD model that systematically 
breaks down complex problems into manageable subquestions, and a QA model that 
accurately answers these subquestions. 
AutoPRM employs a robust training methodology, incorporating supervised 
fine-tuning, feedback-based step-wise verifier, and a final RL 
fine-tuning for the best performance. 
Through extensive experiments, we demonstrate that AutoPRM significantly outperforms 
SOTA methods in terms of efficiency and accuracy on three arithmetic and commonsense reasoning 
tasks. 
Results show that our automated QD process, coupled with a RL-optimized QA model, 
leads to a substantial improvement in handling complex reasoning tasks.

Future developments of AutoPRM could focus on expanding its application to a wider 
range of complex problem domains to test its versatility and identify domain-specific 
challenges.
Additionally, extending its capabilities for long-term reasoning and exploring 
interdisciplinary applications could also be another promising direction.